\documentclass[10pt,twocolumn,letterpaper]{article} 

\usepackage{avss}
\usepackage{times}
\usepackage{epsfig}
\usepackage{graphicx}
\usepackage{amsmath}
\usepackage{amssymb}
\usepackage{fancyhdr}
\usepackage[final]{pdfpages}
\avssfinalcopy % *** Uncomment this line for the final submission

\def\avssPaperID{56} % *** Enter the AVSS Paper ID here

\ifavssfinal\fi
\begin{document}
%%%%%%%%% TITLE
\title{Generative Adversarial Models for People Attribute Recognition in Surveillance}
\author{Matteo Fabbri\\
%University of Modena and Reggio Emilia\\
%via Vivarelli 10 Modena 41125 Italy\\
%{\tt\small matteo.fabbri@unimore.it}
\and
Simone Calderara\\
\\
University of Modena and Reggio Emilia\\
via Vivarelli 10 Modena 41125 Italy\\
{\tt\small name.surname@unimore.it}
\and
Rita Cucchiara\\
%University of Modena and Reggio Emilia\\
%via Vivarelli 10 Modena 41125 Italy\\
%{\tt\small name.surname.it}
}
\maketitle
% \thispagestyle{empty}

%%%%%%%%% BODY TEXT
%%%%%%%%% ABSTRACT
\begin{abstract}
%Video surveillance is an important tool for preventing threats to the public.
%Surveillance cameras has spread rapidly in most of the cities all around the world. Among the surveillance tasks computer vision has focused on tracking and detection of targets where targets are described by their visual appearance. 
In this paper we propose a deep architecture for detecting people attributes (e.g. gender, race, clothing ...) in surveillance contexts. Our proposal explicitly deal with poor resolution and occlusion issues that often occur in surveillance footages by enhancing the images by means of Deep Convolutional Generative Adversarial Networks (DCGAN). Experiments show that by combining both our Generative Reconstruction and Deep Attribute Classification Network we can effectively extract attributes even when resolution is poor and in presence of strong occlusions up to 80\% of the whole person figure.

%Security is of fundamental importance in a world where terrorist attacks are steadily increasing. Governments and agencies face these realities every day, but not always the means at their disposal are sufficient to effectively prevent those attacks. The security area uses many science and engineering fields, and many are the areas of study available. Among the many research opportunities this work addresses the problem of the classification of attributes (such as age, sex, etc.) and items (backpacks, bags, etc.) of people through security cameras. 
%Computer Vision based Deep Learning techniques and generative models are exploited to address this problem in an automatic fashion. The objective of the work is to explore the generalization capability of adversarial networks to enhance people image resolution, typically too low when acquired by surveillance cameras. The network is subsequently exploited to detect and classify people attributes by exploiting the power of convolutional models. Eventually, experiments demonstrate that such an approach can improve state of the art results both when the target resolution is poor and the target image is corrupted or occluded.
\end{abstract}
\section{Introduction}
\begin{figure}[t]
\begin{center}
   \includegraphics[width=.9\columnwidth]{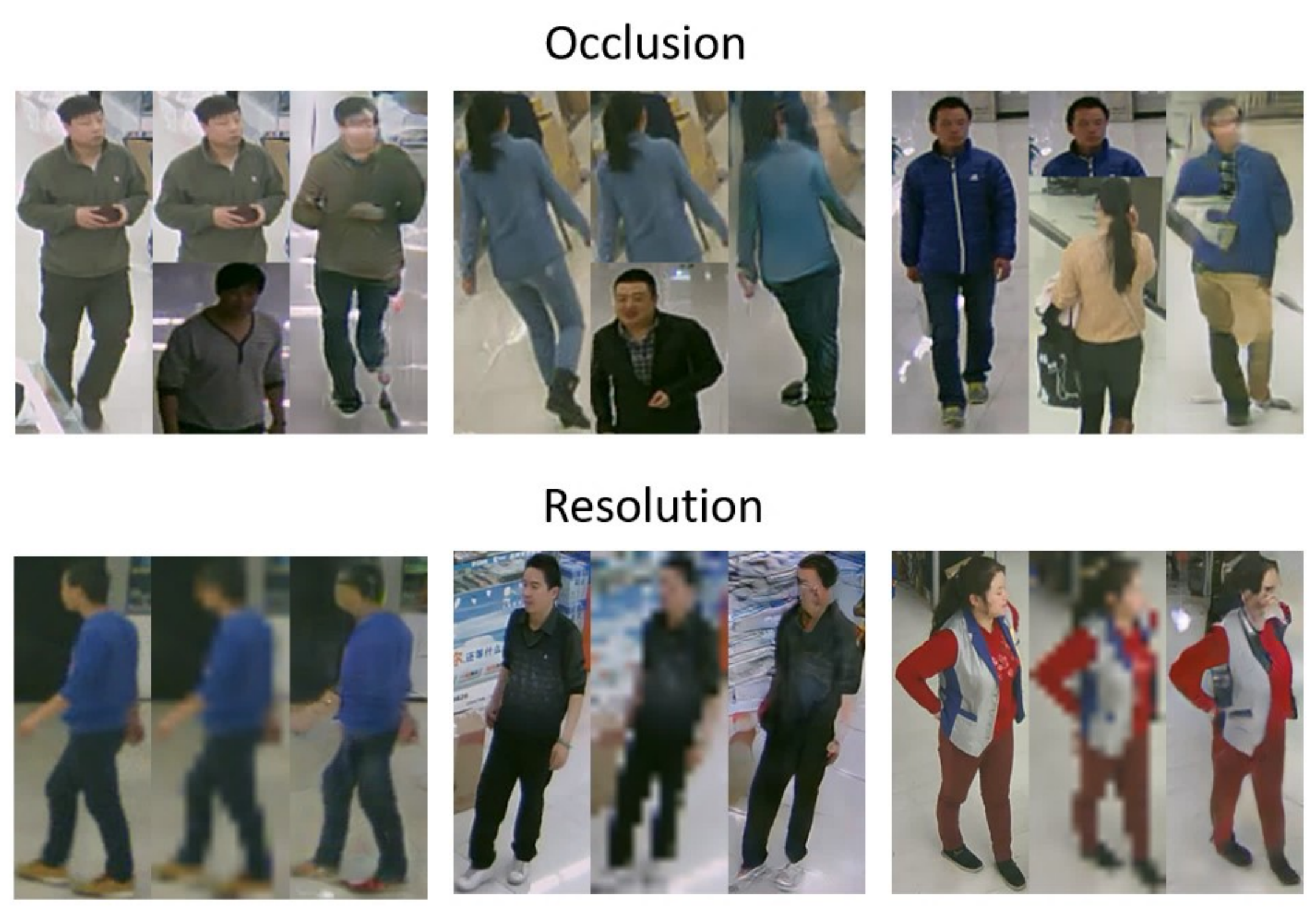}
\end{center}
   \caption{Resolution and occlusions issues and reconstructed frames by our generative approach.}
   \label{fig:1}
\label{fig:front}
\end{figure}

%Security is of fundamental importance in a world where threats are steadily increasing. Governments and agencies face these realities every day, but not always the means at their disposal are sufficient to effectively prevent those attacks. Technology have risen as a valuable help in assessing safety and in particular surveillance cameras are nowadays present in almost all cities in the world.  
%Video surveillance is an important tool for preventing threats to the public.
Surveillance cameras has spread rapidly in most of the cities all around the world. Among the surveillance tasks computer vision has focused on tracking and detection of targets where targets are described by their visual appearance. 
%Surveillance systems capabilities are variegated spanning from detecting people and objects to tracking and high level reasoning.
Nevertheless, recently the task of capturing as many people characteristics as possible has gained importance for better understanding a situation and its attendants. The task, referred in literature as \emph{attribute recognition} \cite{rap}, consists in detecting people attributes (such as age, sex, etc.) and items (backpacks, bags, etc.) of people through security cameras. 
While this task have been profitably attacked from a face recognition perspective capturing gender, age, and race,\cite{other3,other2}, very few works focus on whole people body. Among these, most of them, \cite{other8,pedestrian4}, consider people always unoccluded and at full resolution that is not the case when dealing with surveillance footages. In fact, surveillance cameras, that have typically a far field of view, are massively affected by resolution issues and people occlusion, Fig. \ref{fig:1}. \\In this work we propose an attribute recognition method that explicitly deals with resolution and occlusions by exploiting a generative deep network approach \cite{dcgan}. Our proposal consists of three deep networks. The first classifies people attributes given full body images. The others focus on enhancing the input image by raising its resolution and trying to reconstruct images from occlusion by means of a generative convolutional approach \cite{gan_sr2}. To our knowledge, this is the first work that considers this task in a surveillance context by explicitly dealing with those both issues.

\begin{figure*}
\begin{center}	\includegraphics[scale=0.70]{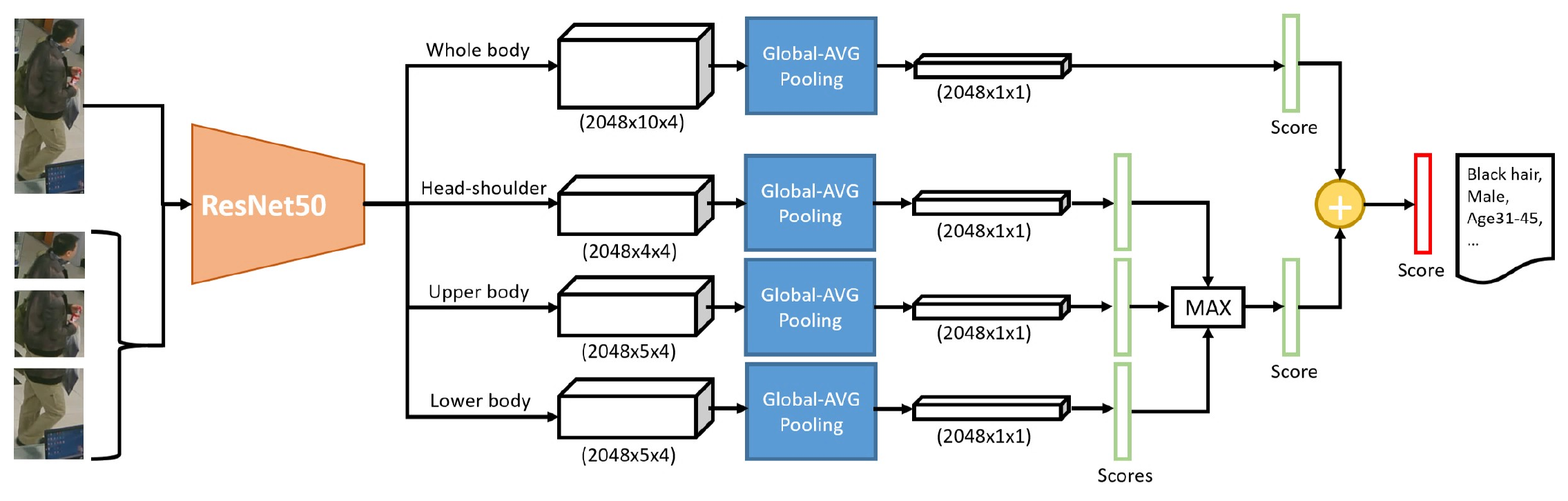}
\end{center}
   \caption{Architecture of our Attribute Classification Network.}
\label{fig:class}
\end{figure*}
%------------------------------------------------------------------------- 
\section{Related Work}
Early works on attribute recognition usually treat attributes by independently training a different classifier for each attribute, \cite{pedestrian1,pedestrian2}. More recently Convolutional Neural Networks (CNN), enable researchers to mine the relationship between attributes and are preferred on large scale object recognition problems because of their advanced performances. There are large bodies of work on CNNs, like \cite{other3} which undertakes the task of occlusion and low-resolution robust facial gender classification, or \cite{other2,other4} that predict facial attributes from faces in the wild. Many other works like \cite{other6,other7} propose different methods to achieve attribute classification like gender, smile and age in an unconstrained environment. However, those technique involve only facial images and are not suitable for surveillance tasks.
Moreover \cite{other8} addresses the problem of describing people based on clothing attributes. Nevertheless, our work considers the person as a whole and does not focus only on clothing classification.
More recent works that rely on full-body images to infer human attributes are the Attribute Convolutional Net (ACN) and Deep learning based Multi-Attribute joint Recognition model (DeepMAR), \cite{acn,pedestrian4}.
ACN jointly learns different attributes through a jointly-trained holistic CNN model, while DeepMAR utilizes the prior knowledge in the object topology for attribute recognition. In \cite{part_cnn1, part_cnn3} attributes classification is accomplished by combining part-based models and deep learning by training pose-normalized CNNs. Additionally, MLCNN \cite{pedestrian3} splits the human body in 15 parts and train a CNN for each of them while DeepMAR* \cite{rap} divides the whole body in three parts which correspond to the headshoulder part, upper body and lower body of a pedestrian respectively. Furthermore, \cite{context} tackles the problem of attribute recognition by improving a part-based method within a deep hierarchical context.
Nevertheless, the majority of those methods relies on high resolution images and does not encompass the problem of occlusion. Recent works on image super-resolution exploit Generative Adversarial Networks (GAN) \cite{gan}, and more precisely Deep Convolutional Adversarial Networks (DCGAN) \cite{dcgan}, in order to generate high resolution images starting from low resolution ones \cite{gan_sr1, gan_sr2}.
% We follow this path and our proposal combines the best of both world by exploiting the power of DCGAN for image enhancement solving occlusion and resolution problems in video surveillance attributes recognition.

%------------------------------------------------------------------------- 
\begin{figure*}
\begin{center}
	\includegraphics[scale=0.54]{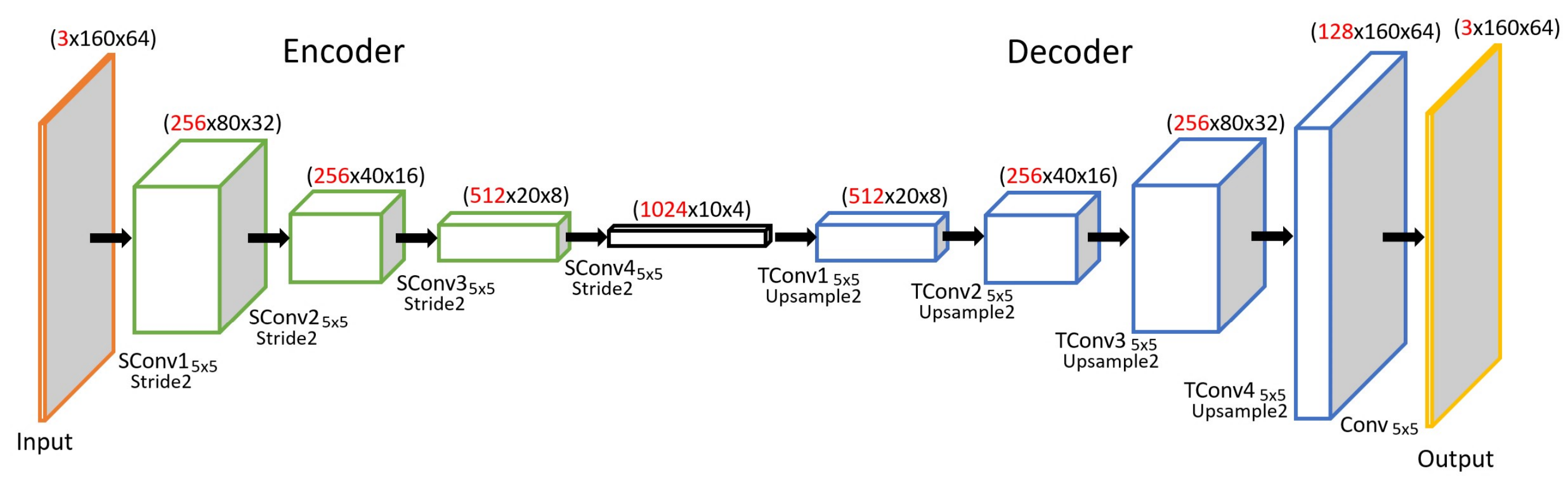}
\end{center}
   \caption{Architecture of our Reconstruction Network.}
\label{fig:gen}
\end{figure*}

\section{Method}
The contribution of this work consists on three networks: a baseline state-of-the-art part-based architecture for human attribute classification based on ResNet \cite{resnet}, a generative model that aims to reconstruct the missing body parts of people in occluded images and a second generative model that is capable of enhance the resolution of images at low resolution.

\subsection{Attribute Classification Network}
The proposed approach for human attribute classification is inspired by the previous part-based works thus capable of learning pose-normalized deep feature representations from different parts. By blending together the capability of neural networks with a part-based approach grants robustness when facing unconstrained images dominated by the effects of pose and viewpoints.
Inspired by \cite{context}, we propose to decompose the input image $I$ into blocks which correspond to the whole body $b$ and a set of parts $\{ p \in P \}$, (inputs in Fig \ref{fig:class}). We choose three parts: head-shoulder section, upper body and lower body of the pedestrian. Those four blocks are then passed through the ResNet50 network \cite{resnet} pretrained on ImageNet classification \cite{imnet} to obtain four part-based convolutional feature maps. Note that, in order to achieve this, we replaced the last average pooling $7 \times 7$ in ResNet50 with a global average pooling. This allowed us to fed the network with images that have one of their dimension smaller than $227$. After computing the feature maps, we branch out two attribute score paths. On the first path, we use the full body feature maps in order to obtain a prediction score based on the whole person. We incorporate the part-based technique in the second path where we compute a prediction score for each image partition (scores in Fig \ref{fig:class}), followed by a $\max$ score operation that aims to select the most descriptive part for each attribute. This operation is needed because human attributes often reside in a specific body area (e.g. "hat" is in the head-shoulder part). The final attribute prediction is performed by adding the whole body score to the part score:
\begin{multline}
\label{eq:eg}
	Score_i(I) = Score_i(b) + Score_i(P) \\
	= w_{i,b}^T \cdot \phi(b) + \max_{p \in P}  w_{i,p}^T \cdot \phi(p)
\end{multline}
where $w_{i,\cdot}$ are the scoring weights of the $i$th attribute for different regions, while $\phi(\cdot)$ are the feature maps from different regions.

The whole network, depicted in Fig. \ref{fig:class}, is trained using a \textit{weighted binary cross entropy} loss with a different weight for each attribute. The need of using a weighted loss arise from the fact that the distributions of attributes classes in the dataset may be imbalanced:
\begin{multline}
\label{eq:eg}
	Loss_{C} = -\sum_{i=1}^A \frac{1}{2 r_i} \cdot y_i \cdot \log( \hat{y_i}) \\
	+ \frac{1}{2(1 - r_i)} \cdot (1 - y_i) \cdot \log(1 - \hat{y_i})
\end{multline}
where A is the total number of attributes, $y$ is the ground truth label vector and $\hat{y}$ is the predicted label vector. For each attribute, $r$ is the ratio between the number of samples that hold that attribute and the total number of samples.
\subsection{Reconstruction Network}
\label{subsec:reconstruction}
Our second goal consist in making the previous architecture robust to occlusion integrating a system capable of removing the obstructions and replacing them with body parts that could likely belong to the occluded person. In the worst case scenario, even though the replaced body parts does not reflect the real attributes of the person, the reconstruction still helps the classifier: by removing the occlusion we produce an image that contains only the subject without noise that could lead to misclassifications. For example, an image containing a person occluded by another person could induce the network to classify the attributes of the person in the foreground which is not the subject of the image.
In order to accomplish this, we train a generative function $G^R$ capable of estimating a reconstructed image $I^R$ from an occluded input image $I^{O}$. During training $I^{O}$ is obtained by artificially partially overlapping an image $I$ with another image. To achieve our goal we train a generator network as a feed-forward CNN $G^R_{{\theta}_g}$ with parameters ${\theta}_g$. For $N$ training images we solve:
\begin{equation}
\label{eq:eg}
	\hat{\theta}_g = \arg \min_{{\theta}_g} \frac{1}{N} \sum_{n=1}^N Loss_R(G^R_{{\theta}_g}(I^{O}_n), I_n)
\end{equation}
Here $\hat{\theta}_g$ is obtained by minimizing the loss function $Loss_{R}$ described at the end of this section.

Following \cite{gan}, we further define a discriminator network $D^R_{{\theta}_d}$ with parameters ${\theta}_d$ that we train alongside with $G^R_{{\theta}_g}$ in order to solve the adversarial min-max problem:
\begin{multline}
\label{eq:eg}
	\min_{G^R} \max_{D^R} \mathbb{E}_{I \sim p_{data}(I)}[\log{D^R(I)}] \\
	+ \mathbb{E}_{I^{O} \sim p_{gen}(I^{O})}[\log{1 - D^R(G^R(I^{O}))}]
\end{multline}
The purpose of the discriminator $D^R$ is to distinguish generated images from real images, meanwhile the generator $G^R$ is trained with the aim of fooling the discriminator $D^R$. With this approach we obtain a generator model capable of learning solution that are similar to not occluded images thus indistinguishable by the discriminator.
Inspired by \cite{dcgan} we propose the generator's architecture illustrated in Figure~\ref{fig:gen}. Specifically, in the encoder we use four strided convolutional layers (with stride 2) to reduce the image resolution each time the number of feature is doubled, SConvs in Fig. \ref{fig:gen}. The decoding uses four transposed convolutional layers (also known as fractionally strided convolutional layers) to increase the resolution each time the number of feature is halved, and a final convolution, TConvs in Fig. \ref{fig:gen}. We use Leaky ReLU as activation function in the encoding phase and ReLU in the decoding phase. We adopt batch-normalization layers before activations (except for the last Conv) and a kernel size $5 \times 5$ at each step.
The discriminator architecture is similar to the generator's encoder except for the number of filter, which increase by a factor of 2 from 128 to 1024. The resulting 1024 feature maps are followed by one sigmoid activation function in order to obtain a probability useful for the classification problem. We use batch-normalization before every Leaky ReLU activation, except for the first layer.

The definition of the loss function $Loss_{R}$ is fundamental for the effectiveness of our generator network. Borrowing the idea from \cite{gan_sr2}, we propose a loss composed by a weighted combination of two components:
\begin{equation}
\label{eq:eg}
	Loss_R = Loss_{SSE} + \lambda Loss_{gen}
\end{equation}
Here $Loss_{SSE}$ is the reconstruction loss based on sum of squared errors of prediction (SSE) which let the generator predict images that are pixel-wise similar to the target image. The pixel-wise SSE is calculated between downsized versions of the generated and target images, first applying an averaged pooling layer. This is because we want to avoid the standard "blurred" effect that MSE and SSE trained autoencoders suffer from. In our experiments we used a $\lambda$ equals to $10^{-1}$.

The second component $Loss_{gen}$ is the actual adversarial loss of the generator $G^R$ which encourages the network to generate perceptually good solutions that are in the subspace of person-like images. The loss is defined as follows:
\begin{equation}
\label{eq:eg}
	Loss_{gen} = \sum_{i=1}^N \log(1 - D^R(G^R(I^{O})))
\end{equation}
Where $D^R(G^R(I^{O}))$ is the probability of the discriminator labeling the generated image $G^R(I^{O})$ as being a real image. 

\subsection{Super Resolution Network}
\label{subsec:multires}
Our last goal is to integrate our system with a network capable of enhancing the quality of images that have poor resolution.
This task is accomplished by training another generative function $G^S$ capable of estimating an high resolution image $I^{H}$ from a low resolution input image $I^{L}$. During training $I^{L}$ is obtained from the original image $I$ by performing a simple downsample operation with factor $r = 4$. To achieve our goal we train the generator network as a feed-forward CNN likewise we did for $G^R$.
As for the Reconstruction Network, we define the discriminator $D^S$ in the same way we defined $D^R$. For $D^S$ we used the same architecture used in $D^R$. 
The architectural differences between the two models reside in the number of layers: in the Super Resolution Network we used three strided convolution (in the Encoder), with 256, 512 and 1024 features respectively and five transposed convolutions (in the Decoder) that follow the pattern 512, 256, 256, 128, 128.  The motivation is that the input image in $G^S$ is two time smaller with respect to the input image in $G^R$.
Moreover in $D^S$ we used five strided convolution with the number of filter that increase by a factor of 2 from 128 to 2048.
%The only architectural difference is between $G^S$ and $G^R$ and it resides in the encoder: in our Super Resolution Network we use only two strided convolution instead of four with 256 and 1024 feature respectively. The reason behind this decision is driven by the fact that the input image in $G^S$ is four time smaller in respect of the input image in $G^R$.
Eventually we set the $\lambda$ value used to weight the loss components to $1$.
%------------------------------------------------------------------------- 
\section{Experiments}
\begin{figure}
\begin{center}
	\includegraphics[width=\columnwidth]{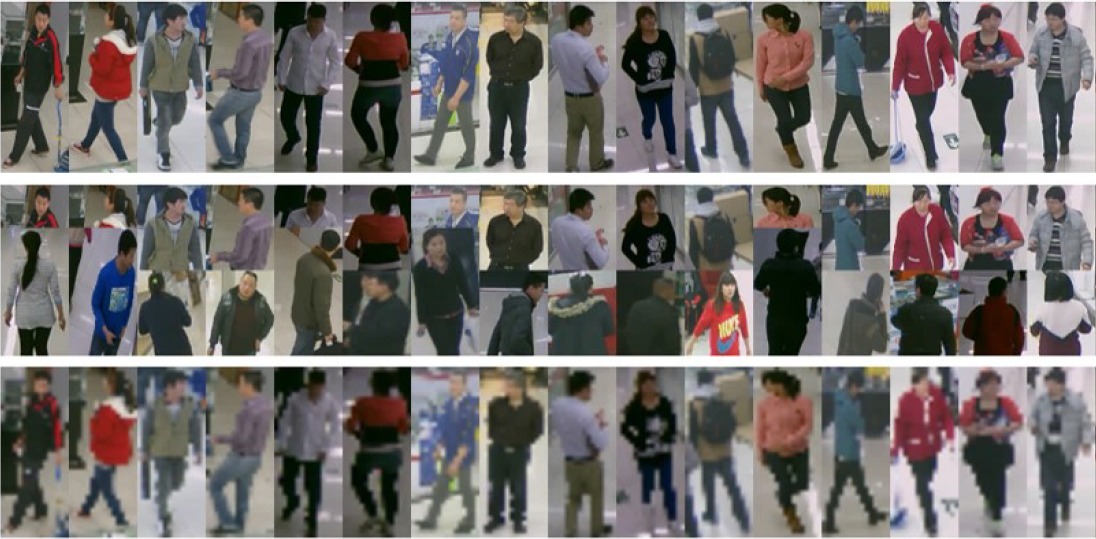}
\end{center}
   \caption{First row original RAP dataset. Second row occlusion RAP dataset where the images have been occluded from 50 to 80\% by pasting the upper body of another subject. Third row low resolution RAP where the image are downsamples of a factor 4 on width and height respectively.}
   %\label{datasets}
\label{fig:rap}
\end{figure}

\begin{figure}
\begin{center}
	\includegraphics[width=\columnwidth]{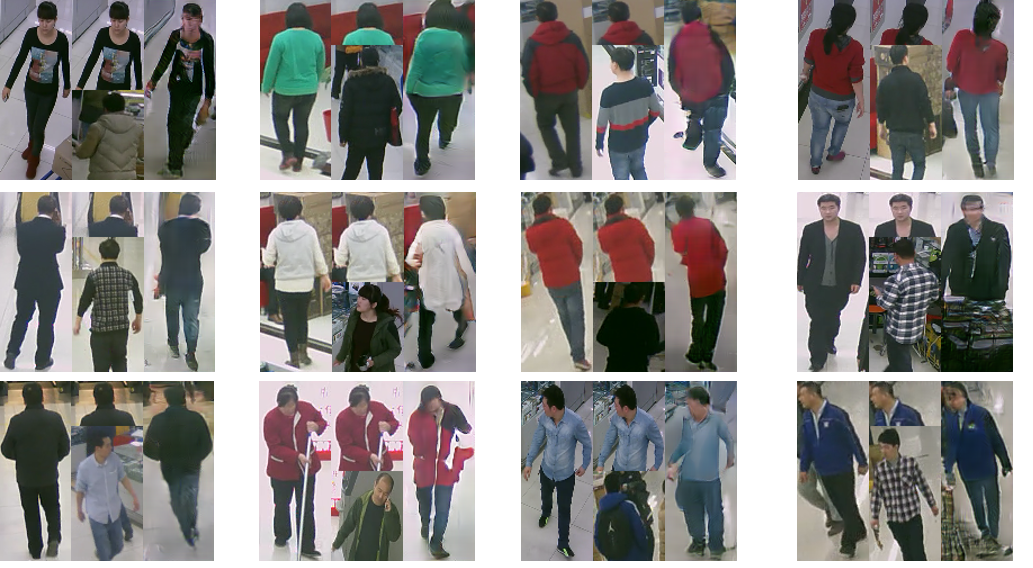}
\end{center}
   \caption{Visual examples of Generative Reconstruction. For every triplet of images: (Leftmost)the original image; (Middle)Occluded input; (Rightmost)Reconstruction/Guessed unoccluded image.}
\label{fig:reco}
\end{figure}

We conduct our experiments using the new RAP \cite{rap} dataset, a very richly annotated dataset with 41,585 pedestrian samples, each of which is annotated with 72 attributes as well as viewpoints, occlusions and body parts information. As recommended by \cite{rap}, we measure performances with five metrics: mean accuracy (mA), accuracy, precision, recall and F1 score.
We perform three type of experiments. First, we comparative evaluated the performances of our Attribute Classification Network comparing the results with other deep SoA approaches. Secondly, we corrupted the dataset with occlusions (occRAP $2^{nd}$ row in Fig. \ref{fig:rap})and tested the benefits achieved by the combination of our Reconstruction Network, Sec. \ref{subsec:reconstruction} with the Attribute Classification one. Thirdly, we corrupt the dataset by lowering the resolution (lowRAP $3^{rd}$ row in Fig \ref{fig:rap}) and evaluate the contribution of our Superesolution Network, Sec. \ref{subsec:multires}, in conjuction with Attribute Classification. Eventually, we propose a complete classification pipeline where all the three network are combined together on both low-res and occluded images.
%We conduct our experiments using the new RAP \cite{rap} dataset, a very richly annotated dataset with 41,585 pedestrian samples, each of which is annotated with 72 attributes as well as viewpoints, occlusions and body parts information. As recommended by \cite{rap}, we measure performances with five metrics: mean accuracy (mA), accuracy, precision, recall and F1 score. Mean accuracy is formally calculated as follows:
%\begin{equation}
%\label{eq:eg}
%	mA = \frac{1}{2A} \sum_{n=1}^A \frac{TP_i}{P_i} + \frac{TN_i}{N_i}
%\end{equation}
%Where, for each attribute $i$, $TP_i$ and $TN_i$ are the number of correctly predicted positive and negative example, while $P_i$ and $N_i$ are the total number of positive and negative example.
The occRAP dataset is produced by randomly overlapping RAP images in order to artificially reproduce the occlusions. Note that in our experiment we focused the attention only on one type of occlusion: the occlusion that cover the bottom part of an image where the occluded portion have been randomly sampled from 50\% to 80\% occlusion rate. lowRap, instead, is obtained by performing a simple downsample with factor 4 from the original RAP images.
\paragraph{Attribute Classification}
Following \cite{rap} we conducted the experiments on the RAP dataset with 5 random splits. For each split, totally 33,268 images are used for training and the rest 8,317 images are used for testing. Due to the unbalanced distribution of attributes in RAP we selected the 50 attributes that have the positive example ratio in the dataset higher than 0.01. For each image we also add one attribute corresponding to the occlusion of our interest (\emph{occlusion down} attribute).
For each mini-batch, we resized the images to a fixed dimension of $320 \times 128$. In order to split the figure in the three parts we divide the height in 10 blocks and pick the top 4 for the head-shoulder part, the third to the seventh for the upper body part and the sixth to the tenth for the lower body part. The network is trained using stochastic gradient descent with a learning rate $10^{-5}$, learning rate decay $10^{-6}$ and momentum $0.9$. We used 8 images per mini-batch.
Tab.~\ref{tab:comparison} shows the results on RAP dataset where our baseline is compared against state-of-the-art methods on the same 51 attributes \footnote{Complete per-attribute results are in the supplementary material.}. 
It can be shown that our network perform favourably in terms of Accuracy being competitive in both Precision and Recall related metrics. This is mainly due to the adoption of the fixed body part partitions of the image that, in case of people images from surveillance cameras, represent a reliable partition of the body in its parts. Additionally, parts scoring maximization allow for selecting the most reliable score for every individual attribute thus increasing the classification accuracy.
\paragraph{Reconstruction}
We trained our Reconstruction Network with the occRAP training set and simultaneously providing the network with the original not occluded images associated to the inputs in order to compute the $Loss_{SSE}$. For optimization we used Adam with $\beta_1 = 0.5$ and learning rate of $0.002$. We alternate updates to the discriminator and generator network with $K = 1$ as recommended in \cite{gan}.\\
Furthermore, the aim is to quantify the impact that occluded images have in the classification task. We firstly fed our classification network with images picked from occRAP testing set obtaining the results reported in Tab.\ref{tab:curruptExp} while visual examples are depicted in Fig. \ref{fig:reco}. Secondly we repeated the experiment manipulating the input images using our Reconstruction Network with the aim of removing the occlusion. In the same table are reported the results that shows a significant improvement. From the results, it emerges that our Reconstruction Network provide a reasonable guess of the occluded person appearance being able of learning from the visible part a potentially useful image completion. 

\begin{table}
\footnotesize
\begin{center}
\begin{tabular}{l|c|c|c|c|c}

Method & mA & Accuracy & Precision & Recall & F1 \\
\hline\hline
ACN \cite{acn}      		& 69.66 & 62.61 & \bfseries 80.12 & 72.26 & 75.98 \\
DeepMAR \cite{pedestrian4} 	& 73.79 & 62.02 & 74.92 & 76.21 & 75.56 \\
DeepMAR* \cite{rap} 		& 74.44 & 63.67 & 76.53 & 77.47 &  77.00 \\
\hline
Our      & \bfseries 79.73 & \bfseries 83.97 & 76.96 & \bfseries 78.72 & \bfseries 77.83 \\
%Our      & \bfseries 79.72 &  \bfseries 83.97 & 76.02 &  \bfseries 78.92 &  \bfseries 77.44 
\hline
\end{tabular}
\end{center}
\caption{Comparison with SoA on the RAP dataset.}
\label{tab:comparison}
\end{table}

\begin{table}
\footnotesize
\begin{center}
\begin{tabular}{l|c|c|c|c|c}

Input & mA & Accuracy & Precision & Recall & F1 \\
\hline\hline
\multicolumn{6}{c}{\bf{occRap experiment}}\\
occRAP & 57.70 & 61.00 & 33.26 & 41.63 & 33.25 \\
\hline
occRAP + NET &\bfseries 68.81 & \bfseries 74.54 & \bfseries 57.29 & \bfseries 58.91 & \bfseries 58.09 \\
\hline
\multicolumn{6}{c}{\bf{lowRap experiment}}\\
lowRAP & 63.80 & 74.51 & 44.47 & 49.56 & 40.37 \\
\hline
lowRAP + NET &\bfseries 76.02 & \bfseries 80.12 & \bfseries 69.56 & \bfseries 73.12 & \bfseries 71.30 \\
\hline
\multicolumn{6}{c}{\bf{Complete Experiment}}\\
\hline\hline
Corrupted &60.68 & 72.75 & 38.67 & 45.47 & 41.80 \\
\hline
Restored & \bfseries 65.82 & \bfseries 76.01 & \bfseries 48.98 & \bfseries 55.50 & \bfseries 52.04 \\
\end{tabular}
\end{center}
\caption{Experiments with occlusions (occRAP experiment), low resolution (lowRAP experiment) and Complete Model. The complete experiment uses the merge of test sets occRAP and lowRAP. \emph{Corrupted} are the score of the Attribute Classification network on plain input data. \emph{Restored} are the results when using our complete pipeline.}
\label{tab:curruptExp}
\end{table}

\paragraph{Resolution}
The dataset used for training the Super Resolution Network is the lowRAP. We trained the network with the original size images associated to the inputs in order to compute the $Loss_{SSE}$. We adopted the same optimizer and the same $K$ value used for the Reconstruiction Network.
In order to evaluate how the low resolution affects the performance on attributes classification, we inputed the lowRAP test set to our Attribute Classification Network, after a 4x bilinear upsampling. Subsequently, we passed the images through our Super Resolution Network before attribute classification. As can be seen in Tab.~\ref{tab:curruptExp} the adoption of our Super Resolution network leads to an important improvement being able to keep more information w.r.t. the upsampling.
\paragraph{Complete Model}
Our final experiment consists in testing all our networks in order to build a system that is able to detect corrupted images and consequently react performing a restoring operation when possible.
To achieve this we propose a simple algorithm where the input is passed through the Super Resolution Network only if the input image is smaller than the network input. The image is then passed through the Classification Network and, if the \emph{occlusion down} attribute is positively triggered (the test F1 score of the \emph{occlusion down} attribute is $>85\%$), the image is passed through the Reconstruction Network. The reconstructed image is finally fed again to the Classification Network to output the final scores. The test were performed on the merge of occRAP and lowRAP test sets.  Tab.~\ref{tab:curruptExp} highlights the improved results obtained by this pipeline w.r.t. our Deep Attribute Classification Network alone.
%------------------------------------------------------------------------- 
\section{Conclusions}
In this work we presented the use of Deep Generative Network for image enhancing in people attributes classification. Our Generative Network have been designed to overcome two common problems in surveillance scenarios, namely people resolution and occlusions. Experiments have shown that jointly enhancing images before feeding them to an attribute classification network can improve the results even when input images are affected by those issues.
In further works we will explore the fusion of the networks in a single end-to-end model that can automatically choose which enhancement network activates by looking at images at test time. We find this line of work can foster research about the problem of attribute classification in surveillance contexts where camera resolution and positioning cannot be neglected.
\section{Acknowledgements}
The work is supported by the Italian MIUR, Ministry of Education, Universities and Research, under the project COSMOS Prin 2015 programme. 
{\small
\bibliographystyle{ieee}
\bibliography{egbib}
}



\section*{Supplementary material}

# Supplementary material for Generative Adversarial Models for People Attribute Recognition in Surveillance

## 1. Per attribute Results

In Tab.1 results of our classification network on the RAP dataset are presented. The table evaluates the per-attribute results on the 51 attributes used for testing and comparison.

## 2. Qualitative Examples of occlusion reconstruction and super resolution

In Fig. 1 we show some qualitative examples of our Reconstruction Network that is capable of guessing the appearance of the occluded subjects. Please note that the appearance is realistic but still influenced by the occluding person. Nevertheless, also elements of the occluded person are kept in the reconstructed image.
In Fig. 2 we show some qualitative examples of our Super Resolution Network. For every triplet the leftmost image is the full resolution input. The center image is the image downsampled 4x and upsampled to the original size. Rightmost image is our super resolution image from our network.

| Attribute | mA | Accuracy | Precision | Recall | F1 |
|---|---|---|---|---|---|
| Age17-30 | 84.58 | 67.50 | 71.40 | 90.08 | 79.66 |
| Age31-45 | 86.07 | 68.47 | 81.16 | 83.10 | 82.12 |
| AgeLess16 | 78.73 | 93.02 | 66.45 | 90.08 | 76.48 |
| Backpack | 71.21 | 91.17 | 80.09 | 69.17 | 74.23 |
| BaldHead | 79.23 | 93.64 | 60.87 | 84.67 | 70.82 |
| BlackHair | 76.75 | 88.88 | 96.87 | 97.31 | 97.09 |
| BodyFat | 81.02 | 73.92 | 64.27 | 74.96 | 69.20 |
| BodyNormal | 69.59 | 63.01 | 88.12 | 79.37 | 83.52 |
| BodyThin | 74.36 | 78.78 | 85.74 | 63.16 | 72.74 |
| Boots | 78.69 | 87.68 | 82.43 | 68.04 | 74.55 |
| Box | 85.07 | 89.44 | 62.39 | 82.52 | 71.06 |
| Calling | 74.90 | 88.96 | 71.81 | 83.35 | 77.15 |
| CarrybyArm | 73.75 | 90.70 | 69.84 | 91.37 | 79.17 |
| CarrybyHand | 71.15 | 82.12 | 78.60 | 76.04 | 77.30 |
| Casual shoes | 80.81 | 68.95 | 64.29 | 84.64 | 73.08 |
| Clerk | 87.68 | 91.13 | 70.12 | 85.83 | 77.19 |
| Cloth shoes | 84.73 | 92.13 | 74.39 | 78.76 | 76.51 |
| Cotton | 78.59 | 77.12 | 72.95 | 81.19 | 76.85 |
| Customer | 85.84 | 90.52 | 98.68 | 97.24 | 97.96 |
| Dress | 83.62 | 84.34 | 89.91 | 70.83 | 79.24 |
| Female | 87.03 | 84.21 | 88.33 | 90.84 | 89.57 |
| Gathering | 82.05 | 76.39 | 77.20 | 67.20 | 71.85 |
| Glasses | 85.29 | 83.75 | 70.07 | 63.16 | 66.44 |
| HandBag | 85.77 | 90.56 | 66.76 | 85.04 | 74.80 |
| HandTrunk | 77.91 | 91.80 | 79.21 | 88.51 | 83.60 |
| Hat | 80.08 | 91.38 | 63.33 | 62.48 | 62.90 |
| Holding | 79.32 | 90.90 | 63.39 | 90.31 | 74.49 |
| Jacket | 71.91 | 72.29 | 83.30 | 78.25 | 80.70 |
| Jeans | 85.22 | 84.04 | 86.29 | 83.37 | 84.81 |
| Leather Shoes | 76.41 | 77.29 | 70.98 | 81.75 | 75.98 |
| LongTrousers | 81.08 | 78.10 | 95.46 | 89.78 | 92.53 |
| Muffler | 76.42 | 92.43 | 66.95 | 72.52 | 69.62 |
| Other Attachment | 81.57 | 68.59 | 70.41 | 81.06 | 75.36 |
| Other Trousers | 84.48 | 85.47 | 76.91 | 83.39 | 80.02 |
| PaperBag | 74.89 | 91.65 | 79.94 | 65.09 | 71.75 |
| PlasticBag | 80.20 | 90.21 | 81.18 | 64.97 | 72.18 |
| Pulling | 83.10 | 92.20 | 64.82 | 71.90 | 68.18 |
| Pusing | 83.07 | 92.59 | 84.05 | 74.52 | 79.00 |
| Shirt | 77.71 | 76.05 | 85.06 | 79.51 | 82.19 |
| ShortSkirt | 82.62 | 82.08 | 85.88 | 67.43 | 75.4 |
| ShortSleeve | 89.34 | 91.58 | 86.23 | 64.14 | 73.56 |
| Skirt | 82.12 | 83.12 | 88.52 | 68.52 | 77.25 |
| Sport Shoes | 70.15 | 73.20 | 80.41 | 69.28 | 74.43 |
| SSBag | 87.46 | 83.27 | 68.33 | 69.88 | 69.10 |
| SuitUp | 74.41 | 91.16 | 62.35 | 78.52 | 69.51 |
| Sweater | 78.76 | 78.92 | 81.49 | 90.87 | 85.92 |
| Talking | 74.90 | 86.50 | 76.16 | 80.43 | 78.24 |
| Tight | 82.59 | 88.66 | 79.44 | 76.75 | 78.07 |
| TightTrousers | 76.27 | 87.95 | 82.36 | 86.88 | 84.56 |
| TShirt | 70.47 | 75.29 | 72.73 | 67.61 | 70.08 |
| Vest | 77.21 | 89.12 | 76.97 | 89.09 | 82.59 |
| Occlusion Down | 93.21 | 94.18 | 79.31 | 91.68 | 85.04 |

Table 1. Metrics for each attribute.

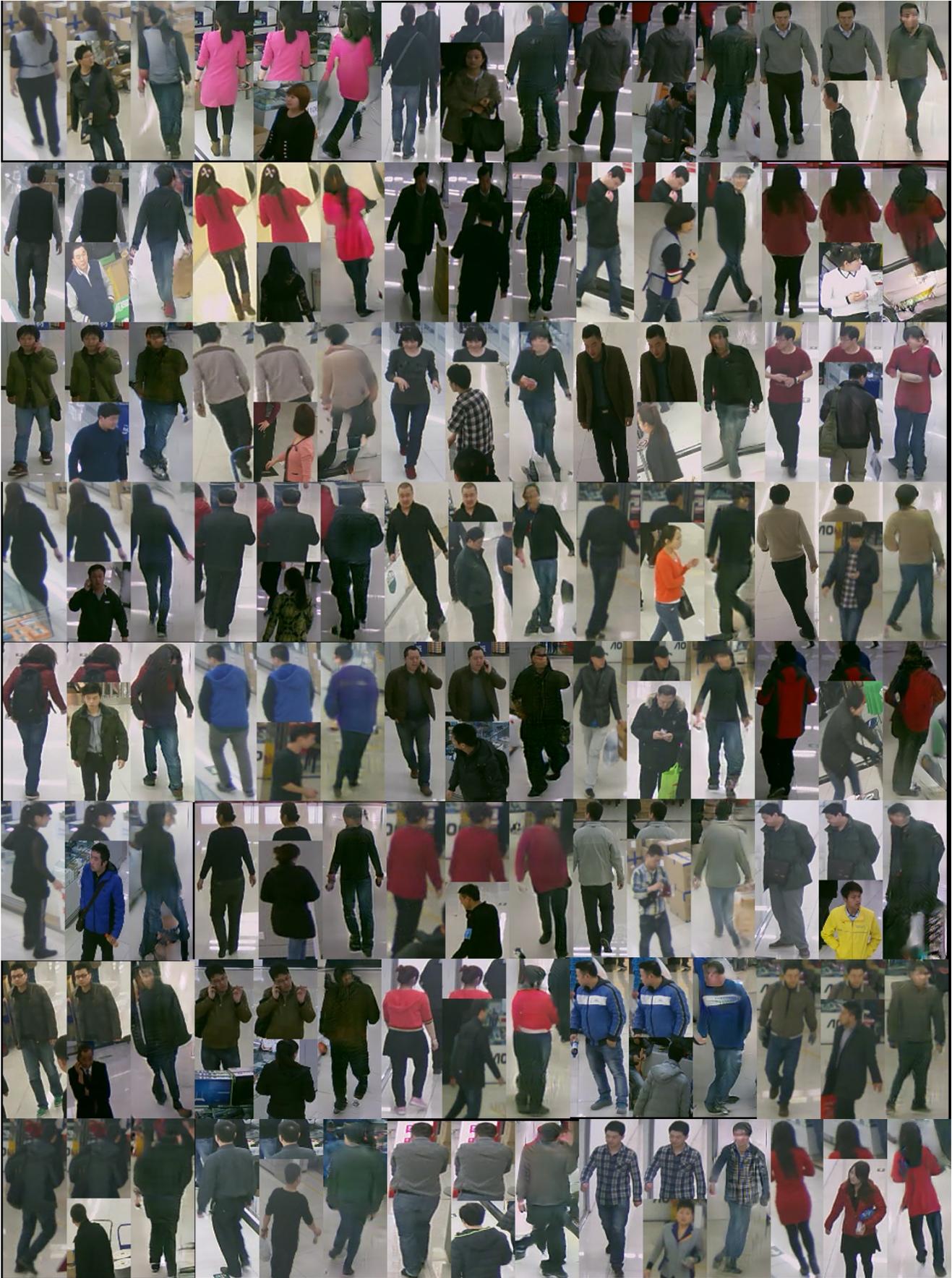

Figure 1. Examples of the output from Reconstruction Network. For every triplet the first image is the ground truth. Middle image is the occluded input to the network. Rightmost image is the unoccluded reconstruction by our Reconstruction Network.

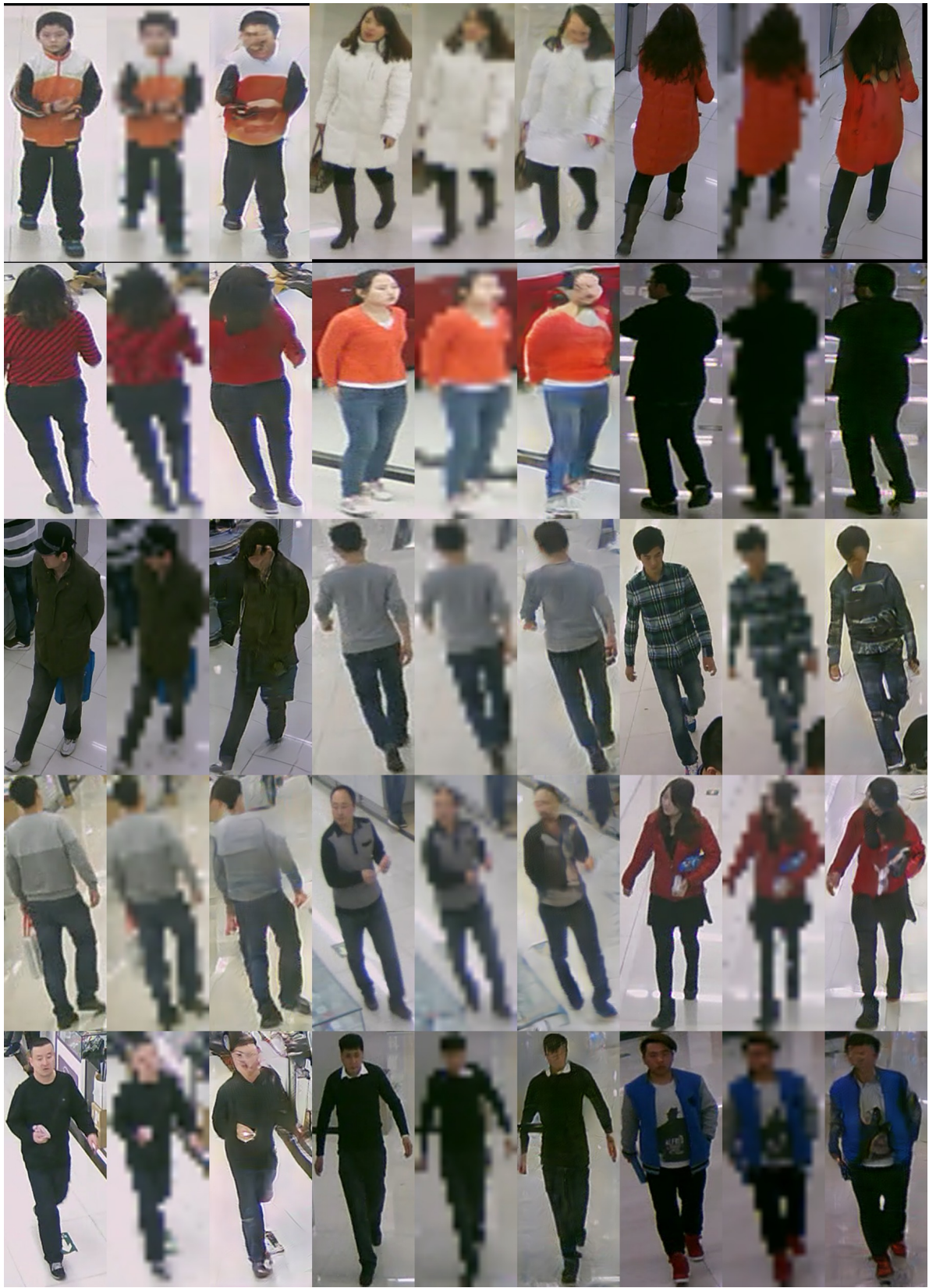

Figure 2. Examples of the output from our Super Resolution Network. For every triplet the first image is the ground truth. Middle image is the downsampled 4x image then upsampled to full size. Rightmost image is the super-resolution image obtained by our deep Super Resolution Network.



\end{document}